\documentclass[letterpaper, 10 pt, conference]{styles/ieeeconf}  
\IEEEoverridecommandlockouts                              %
\IEEEoverridecommandlockouts                              

\usepackage{graphics}
\usepackage{epsfig}
\usepackage{mathptmx}
\usepackage{times}
\usepackage{amsmath}
\usepackage{amssymb}
\usepackage{booktabs}
\usepackage{hyperref}

\title{\LARGE \bf
AARK: An Open Toolkit for Autonomous Racing Research
}

\author{James Bockman$^{1}$
   Matthew Howe$^{1}$
   Adrian Orenstein$^{2}$
   Feras Dayoub$^{1}$
\thanks{$^{1}$James Bockman, Matthew Howe and Feras Dayoub are with The Australian Institute for Machine Learning (AIML), The University of Adelaide, Adelaide, 5000 SA, Australia
        {\tt\small \{james.bockman, matthew.howe,  feras.dayoub\}@adelaide.edu.au}}%
\thanks{$^{2}$Adrian Orenstein is is with The Alberta Machine Intelligence Institute (Amii), The University of Alberta, Edmonton, T6G 2R3 AB, Canada
        {\tt\small aorenste@ualberta.ca}}%
}

\begin{document}

\maketitle
\thispagestyle{empty}
\pagestyle{empty}

\begin{abstract}
    Autonomous racing demands safe control of vehicles at their physical limits for extended periods of time, providing insights into advanced vehicle safety systems which increasingly rely on intervention provided by vehicle autonomy.
    Participation in this field carries with it a high barrier to entry.
    Physical platforms and their associated sensor suites require large capital outlays before any demonstrable progress can be made.
    Simulators allow researchers to develop soft autonomous systems without purchasing a platform.
    However, currently available simulators lack visual and dynamic fidelity, can be expensive to license, lack customisation, and are difficult to use.
    AARK provides four packages, ACI, ACDG, ACMPC and ACRL.
    These packages enable research into autonomous control systems in the demanding environment of racing to bring more people into the field and improve reproducibility: ACI provides researchers with a computer vision-friendly interface to Assetto Corsa for convenient comparison and evaluation of autonomous control solutions; ACDG enables generation of depth, normal and semantic segmentation data for training computer vision models to use in perception systems; ACMPC and ACRL provide full stack autonomous, and reinforcement learning based solutions capable of controlling vehicles to build from.
    AARK aims to unify and democratise research into a field critical to providing safer roads and trusted autonomous systems.
\end{abstract}

\section{INTRODUCTION}
    When vehicles are operated at the edges of their physical limitations a balance must be struck between operator and platform to ensure a favourable outcome; the slightest misstep on either part leads to catastrophic outcomes for both.
    This place, where a platform is pushed to its absolute capability, is where vehicle operators find themselves before incidents that result in damage, injury, and loss of life.
    It is therefore imperative to explore vehicle behaviour in these situations to improve road safety.
    The use of automated assistive systems to bring vehicles back from the brink, in-spite of its operator, are becoming increasingly common.
    Traction control, electronic stability control, and anti-lock braking are all automatic systems that respond to and remedy a vehicle's rapid approach toward or surpassing its capabilities.
    These systems rely on making decisions in response to intrinsic vehicle sensors, in a limited form of autonomy that usurps control from the operator.
    Advanced Driver Assistance Systems (ADAS) takes this further.
    ADAS exists in partnership with the operator, using extrinsic sensors to respond to the environment external to the vehicle.
    This manifests as adaptive cruise control, lane following, blind spot monitoring and automatic braking.
    In this way, manufacturers have been expanding the level of autonomy present in vehicles to become trusted partners of drivers.
    At the other end of the spectrum, fully autonomous systems, capable of controlling vehicles on their own, can be used to provide insights into how current assistive offerings can be expanded further to provide safer roads.
    However, in the classical self-driving problem situations where the vehicle is exerting safety critical control at the edge of platform capability are rare.
    These events are exactly what occur before many road incidents.
    When their physical limits are pushed vehicles behave drastically different from when they are use under normal operation.
    Inputs become detached to their outputs, causing road users to make inputs that push the vehicle further out of control, due to misaligned expectations.
    Without validation of autonomy's capacity to deal with this paradigm shift in control dynamics, how can they be trusted to safely negotiate and recover vehicles from the brink of disaster?
    Traditionally, motorsport has been used to provide an environment that forces vehicles to be operated at the limits of their capability, revealing durability, stability and performance.
    In such events, for a driver to be competitive they must continually flirt with and even cross the barrier between being in and out of control.
    A similar statement can be made when replacing drivers, in such events, with autonomous systems.
    This expands the concerns of traditional self-driving autonomy by providing an imperative that ensures agents operate vehicles at the limits of physical capability.
    Control solutions that are able to race effectively and safely will have an implicit ability to exert corrective control in situations where human operators find themselves prior to disaster.

\begin{figure}
    \centering
    \includegraphics[width=0.46\textwidth]{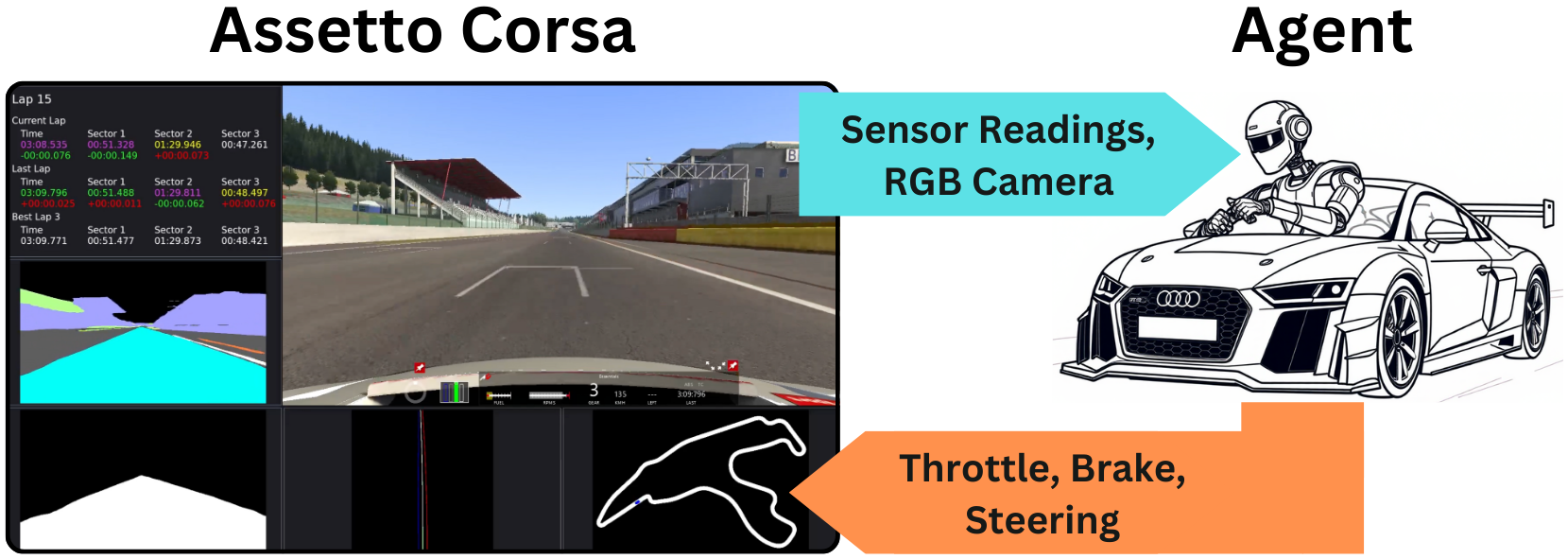}
    \caption{Diagram of information flowing through ACI to enable autonomous systems to drive vehicles in Assetto Corsa}
    \label{fig:aci}
\end{figure}

\section{RELATED WORK}
Previous efforts to provide a simulator for autonomous racing have failed to proliferate into consistent use across research institutions; preventing the unification of the field and limiting reproducibility.
    Preventing factors are specific to each simulator.
    In general, they have been complicated to use, short on customisation options, closed source, or lack visual and dynamic fidelity.
    A series of pedagogical platforms have been developed to support educational programs such as AWS DeepRacer~\cite{deepracer} and F1TENTH~\cite{f1tenth, f1tenthdev}.
    Despite being easy to use, these platforms lack modelling rigour and limit customisation of vehicles.
    A similar deviation from reality in their approach to modelling is present in Open AI Gym's Car Racing~\cite{openai_gym} and Track Mania~\cite{trackmania}.
    Collectively, these simulations use RC scale or imagined vehicles without veracious dynamics modelling.
    Similar deficits in simulator fidelity are also present in full-scale vehicle simulators.
    Car Learning to Act (CARLA)~\cite{carla}, Generalized RAcing Intelligence Competition (GRAIC)~\cite{GRAICrace, GRAIC-CI-OCAR21}, Autoware Simulator (AWSIM) and Learn-to-Race (L2R) are all simulators that have been used in full-scale autonomous racing challenges.
    All of these examples are modified versions of simulators built primarily for the evaluation of driverless road vehicles.
    Although autonomous racing and driving appear to be analogues of one another, there are a series of simplifications that can be applied to dynamics simulation when vehicles are intended to be driven well within their physical limits.
    Self-driving vehicle simulators neglect high fidelity simulation of aerodynamics, fuel usage, tyre degradation and suspension geometry.
    This is because these have a negligible effect for road vehicles driven under normal conditions.
    Additionally, some of these simulators provide visually sparse environments that limit the application of computer vision techniques as part of an autonomous control stack.
    
    In contrast, Assetto Corsa (AC) is a customisable, racing-specific simulator that has a large collection of virtualised world-renowned race tracks and cars rendered to a high level of visual quality.
    AC's inherent customisablity allows it to be used to simulate existing real-world autonomous racing leagues: Indy Autonomous Challenge~\cite{iac}, Autonomous Karting Series~\cite{aks}, Roborace~\cite{roborace} and A2RL~\cite{a2rl}.
    Although the simulator itself is closed source, a mature modification community exists which have exposed internal methods of customisation and developed a range of additional assets such as vehicles, circuits, and visual improvements.
    AC is also regularly on sale for \$4USD~\cite{acprice} making it significantly more accessible compared to proprietary simulators used in motorsport.
    The desirability of AC has also been validated by its use in other work.
    In ~\cite{Raji_2024} there is a section dedicated to validation of their control solution using AC and in the future, an interface would be published openly; this has yet to happen.
    Most recently~\cite{remonda2024simulationbenchmarkautonomousracing}, developed concurrently to AARK, details a similar interface for AC.
    Their interface was not released with the capability to: provide agents with visual sensor information or simulated INS readings, generate machine vision data, run on Unix systems and do not provide a modular full stack autonomous control solution.
    
    Enter the Adelaide Autonomous Racing Kit (AARK); an open source toolkit that enables research, development and validation of autonomous systems capable of operating vehicles pushed to their limits.
    AARK represents a feature rich, extensible manifestation of autonomous racing researchers' desire to validate and compare autonomous systems in AC.
    Built around AC, AARK offers researchers a platform capable of generating data to train machine learning systems in addition to validating and comparing autonomous systems via a series of packages:
    \begin{itemize}
    \item{AC Interface (ACI) - A harness that enables recording agent actions (human \& autonomous), and enables autonomous systems to receive observations from and control vehicles in AC}
    \item{AC Data Generation (ACDG) - A machine learning data generation pipeline that produces depth, normal and semantic annotations from recordings}
    \item{AC Model Predictive Controller (ACMPC) - A full-stack autonomous control solution for use as a prototypical autonomous system}
    \item{AC Reinforcement Learning (ACRL) - A gym environment for training reinforcement learning agents}
    \end{itemize}
    The source code to each of these packages is available on Github \url{https://github.com/orgs/Adelaide-Autonomous-Racing-Kit/repositories} with accompanying documentation \url{https://www.adelaideautonomous.racing/}.

\section{ADELAIDE AUTONOMOUS RACING KIT}

The Adelaide Autonomous Racing Kit (AARK) positions itself in the \textit{no man's land} between two separate philosophical approaches to the autonomous racing problem; classical control and reinforcement learning (RL).
Classical approaches to autonomous racing systems are built up from modular components that each handle specific tasks.
This leads to systems that require laborious hand tuning of non-linear optimisation parameters which are specific to each particular vehicle and circuit~\cite{DBLP:journals/corr/abs-2106-04094, Betz_2023, Raji_2024, jung2023}.
Conversely, RL approaches ingest environmental state and directly produce control outputs.
In the current literature, classical autonomous systems heavily rely on known pose, GPS, LiDAR, and vehicle dynamics modelling.
By contrast, previous RL methods access oracle knowledge of the simulation's internal state~\cite{DBLP:journals/corr/abs-2103-14666, gtsophy, DBLP:journals/corr/abs-2104-11106, remonda2024simulationbenchmarkautonomousracing} and have the luxury of acting synchronously with their environment~\cite{DBLP:journals/corr/abs-1807-02371, deepracer}.
For RL approaches to become practically useful, they need to evolve from relying on perfect state information and account for the fundamentally asynchronous nature of reality.
The strengths of both approaches have the potential to address the weaknesses of the other.
In fact, machine learning approaches are being integrated into classical autonomous system's modules~\cite{AR-review}, albeit slowly, over time.
For these fields to unify and develop together they require a common environment that is equally convenient for researchers of both methodologies to use.
AARK has the potential to bridge this gap by providing a development and evaluation platform for autonomous racing systems that cater to both of these disciplines.

\section{ASSETTO CORSA INTERFACE}
 At its core AARK provides an interface between the popular racing simulator AC and autonomous agents, Figure \ref{fig:aci}.
    The enabling package, ACI, achieves this by scraping vehicle state from the simulator's memory, capturing frames displayed using FFMPEG and relaying control inputs to AC via a virtual controller.
    At a frequency of \~100Hz, ACI outputs 190 different readings describing the simulated vehicle's current state.
    These readings contain detailed information regarding the vehicle's ground truth state, intended for evaluation and debugging.
    Accompanying this are 17 readings from a simulated Inertial Navigation System (INS), intended to be passed to the autonomous driver.
    The INS provides accelerometer, gyroscope, odometer and GPS readings with artificial noise and drift to better reflect real sensors.
    With these readings, ACI also provides image frames captured of the simulator's application window.
    Frame captures are non-blocking, allowing systems to continue performing operations that depend on sensor readings while waiting for new visual input.
    The frequency of frame captures is 30Hz by default but can be configured by the user to suite their requirements and computer's performance.
    Once an observation has been processed by the autonomous system it returns $[\text{throttle}, \text{brake}, \text{steering}]$ to ACI which represents the inputs that should be applied to the vehicle.
    
    Additionally, ACI provides two methods to capture driver data for analysis or training.
    Leveraging an AC plugin, AC Telemetry Interface (ACTI), vehicle performance data can be captured and written out in the prolific MoTeC format.
    MoTeC data is currently used to analyse both vehicle and driver performance across motorsport, providing access to a suite of ready-made visualisation tools to verify system behaviour.
    Vast libraries of driver MoTeC data exist online, for a wide range of operator proficiencies, making it an excellent way to compare autonomous and human driver behaviors.
    However, MoTeC does not offer a convenient way to package visual information synchronised with state information in the necessary form for training AI-based approaches.
    ACI addresses this deficit by recording sessions as a series of image, state pairs at a frequency of ~30Hz.
    This format of recording can be used to train machine vision systems with, either directly or indirectly via post-processed data produced by AC Data Generator (ACDG).
    More details on the implementation framework for ACI are given in appendix \ref{app:aci_implementation}.

\begin{figure}
    \centering
    \resizebox{0.45\textwidth}{!}{
        \begin{tabular}{cc}
            \includegraphics[width=\textwidth]{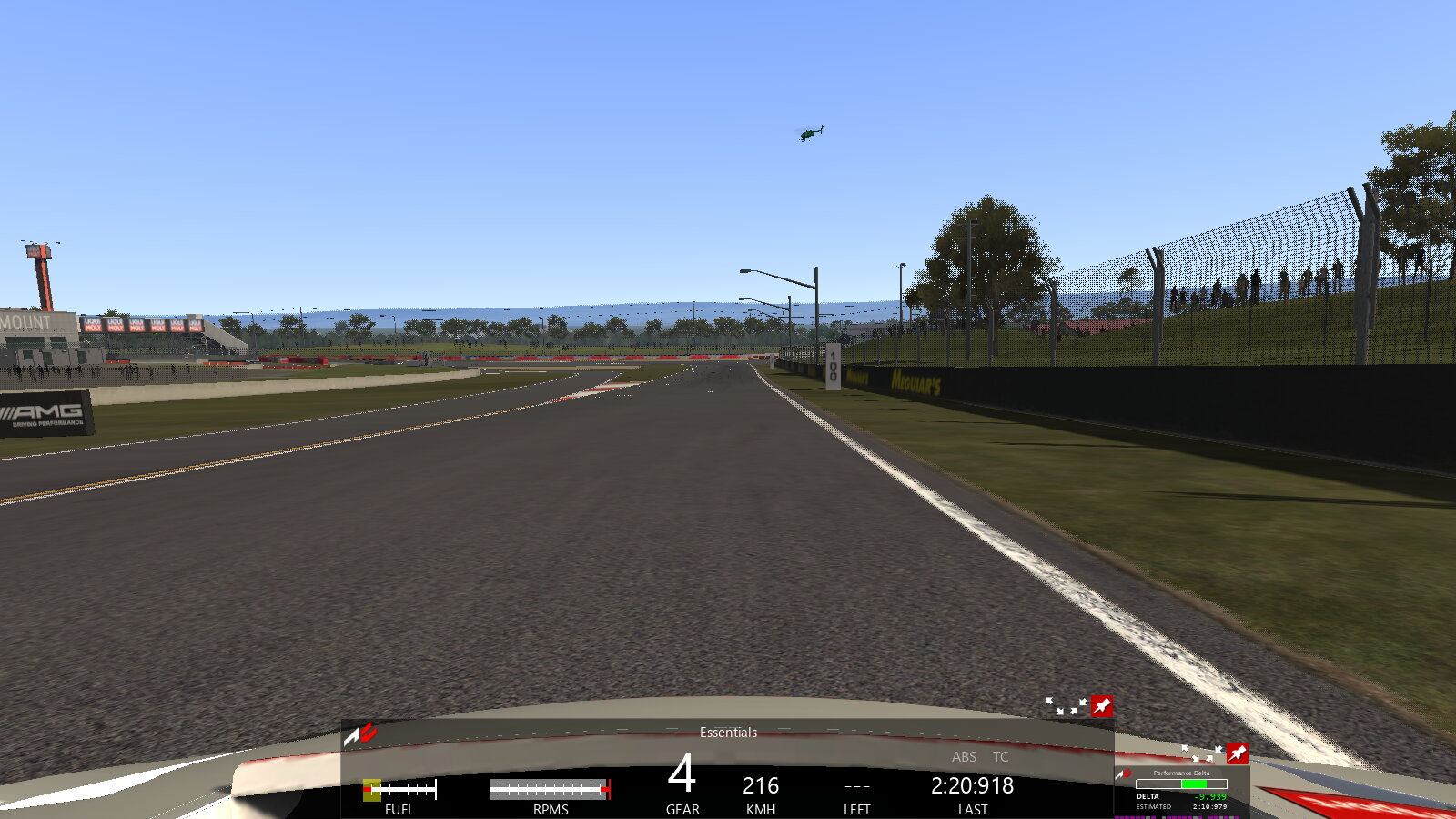}&
            \includegraphics[width=\textwidth]{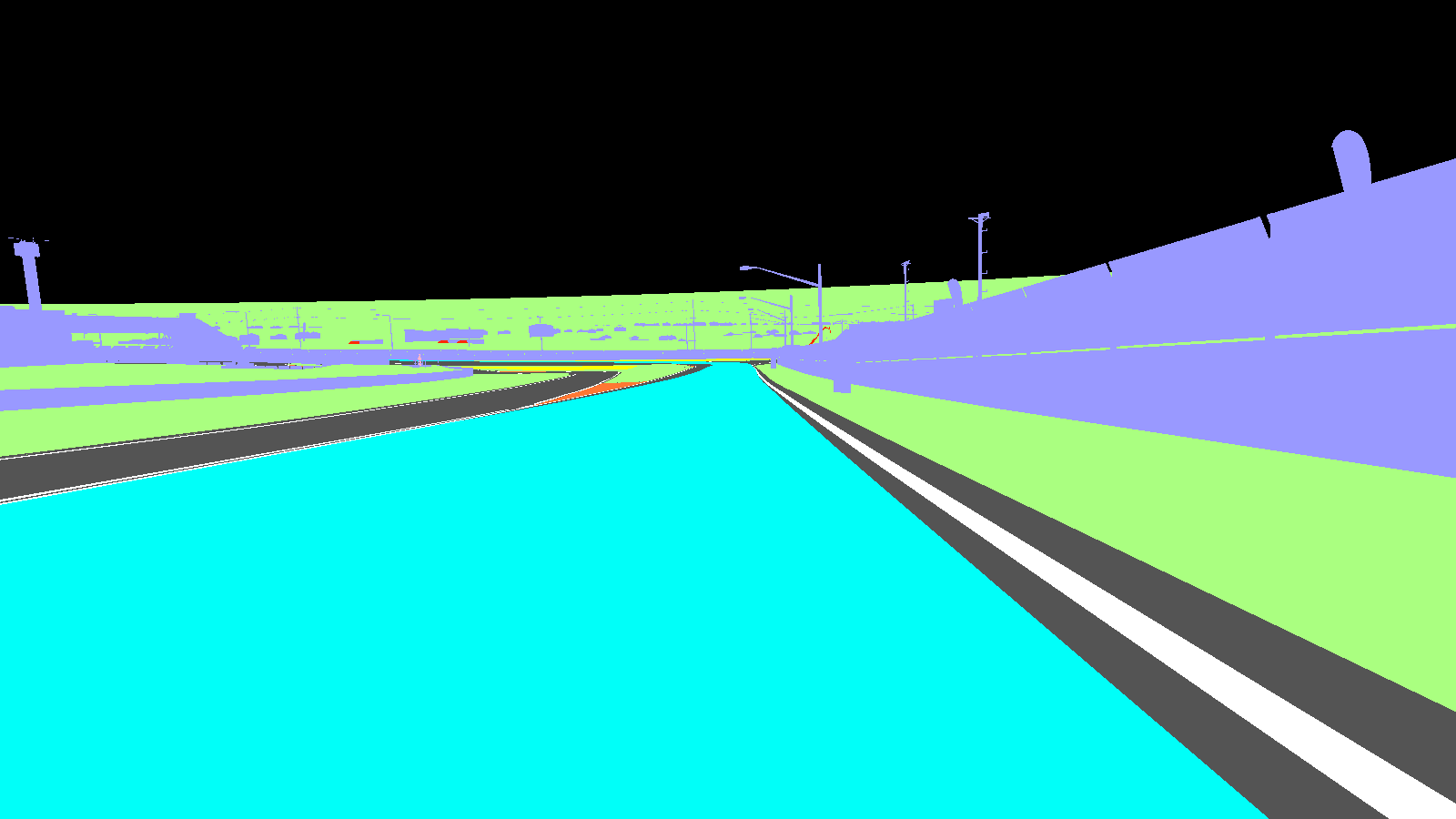} \\
            \includegraphics[width=\textwidth]{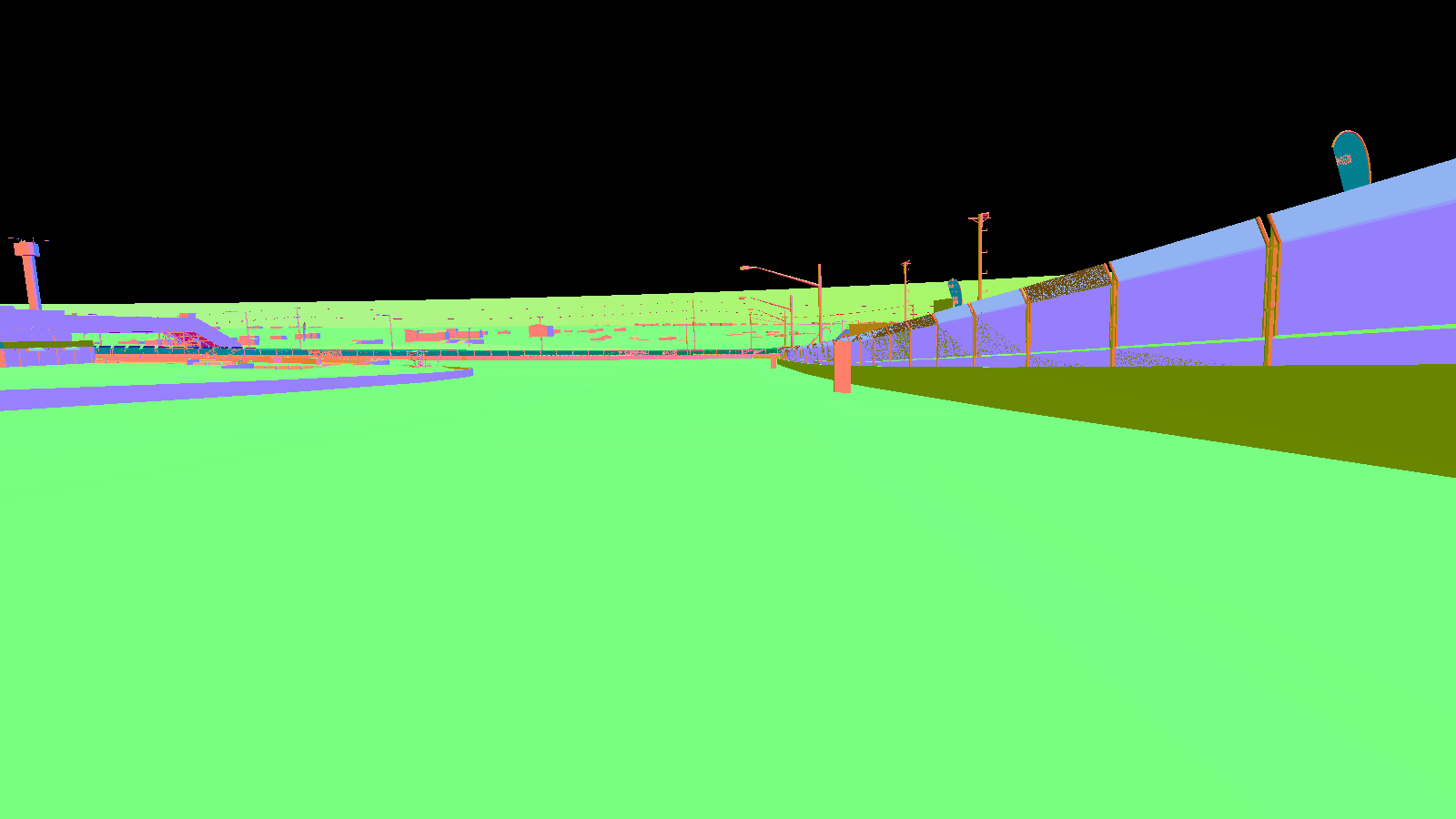}&
            \includegraphics[width=\textwidth]{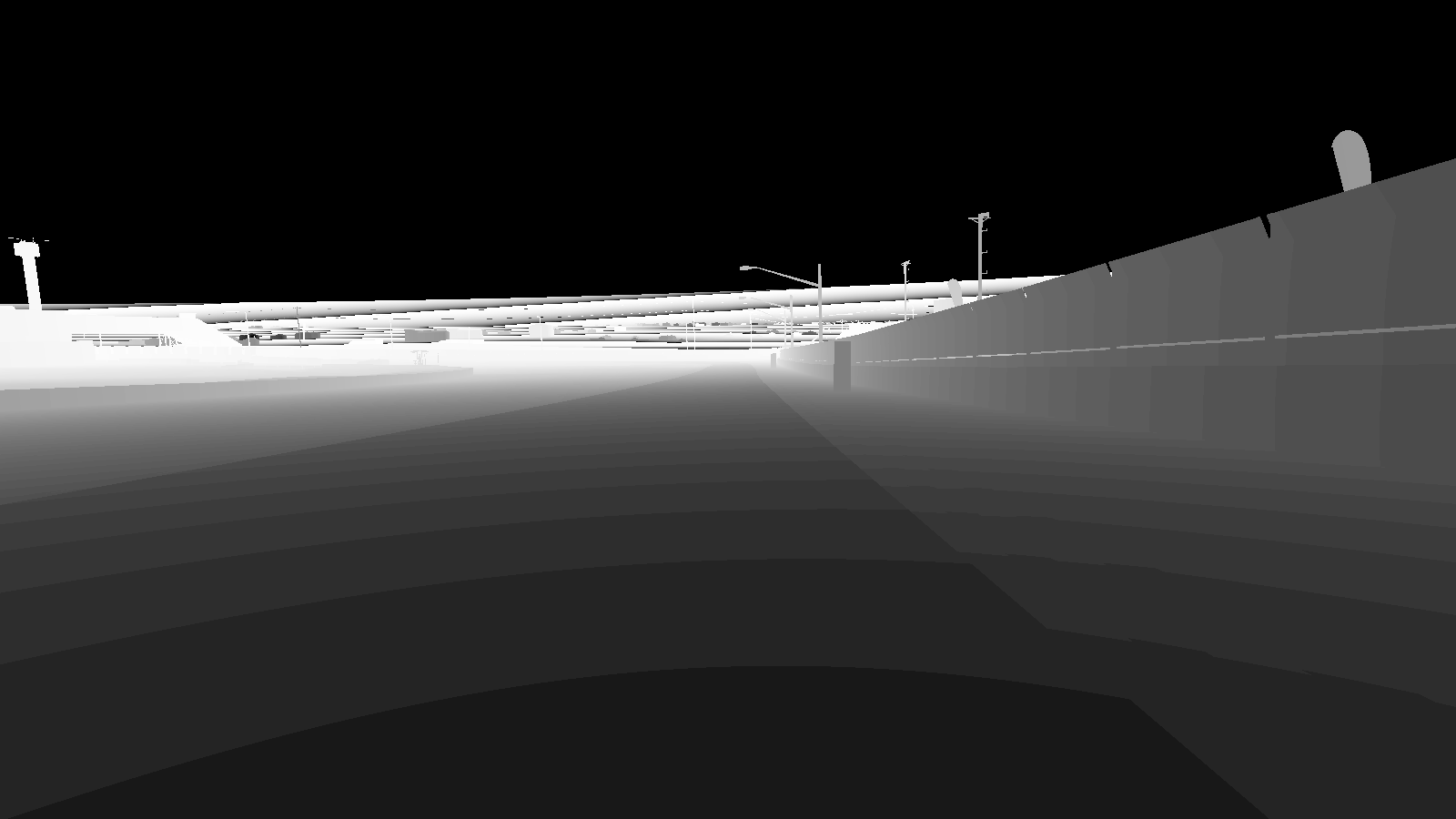}
        \end{tabular}
    }
    \caption{Examples of visualised data generated by our toolkit. From top-left to bottom-right: camera image, semantic segmentation, surface normal and depth maps are shown.}
    \label{fig:data-generation}
\end{figure}

\begin{table}
\centering
\caption{
Details of the semantic segmentation datasets created using ACI and ACDG, in addition to a FPN-Resnet-18 semantic segmentation model's achieved mIoU.
}
\label{tab:segmentation-data}
\begin{tabular}{lccc}
\toprule
Track       &   \#Training & \#Validation & mIoU\% \\ \midrule
Monza  &  15,778  &  2,157 &  86.19 / 85.12\\
Mount Panorama & 24,795 & 1,334 & 82.59 / 82.70 \\
Nordschleife & 62,888 & 5,581 & 84.69 / 82.83 \\ 
Silverstone & 23,023 & 1,296 & 76.14 / 73.65 \\
Spa    &  14,797  & 1,330 &  79.37 / 79.29\\
Vallelunga & 27,878 & 1,018 & 75.44 / 74.49 \\
Yas Marina & 55,602 & 1,630 & 70.84 / 70.89 \\ \bottomrule
\end{tabular}
\end{table}

\section{DATA GENERATOR}
    Machine learning has demonstrated efficacy in tasks that estimate the properties of objects from images of those objects.
    Many of these properties can be leveraged to provide information to an autonomous system about its environment, such as depth, surface normal vectors and semantics.
    These sources of information have been under utilised in previous approaches that rely heavily on LiDAR, raw camera feeds, or ego state.
    The barrier, and likely reason for their lack of proliferation, to applying machine learning based vision techniques is their requirement for a large body of domain specific annotated data.
    For example, to train a semantic segmentation model for use in autonomous racing, video data from each different race track needs to be gathered and human annotators need to specify a label for each pixel in each image corresponding to a racing specific semantic class.
    ACDG automatically generates multiple types of annotated machine learning data from recordings gathered by ACI.
    Figure \ref{fig:data-generation} shows an example of semantic, depth, and surface normal annotations generated for an image by ACDG.
    This is achieved by; recreating the recorded camera perspective at each time step, casting rays, and extracting information about where those rays intersect with the circuit's mesh.
    Depth and surface normal annotations can be generated for any circuit.
     
\begin{figure*}
    \centering
    \resizebox{\textwidth}{!}{
        \begin{tabular}{c}
        \includegraphics[width=0.9\textwidth]{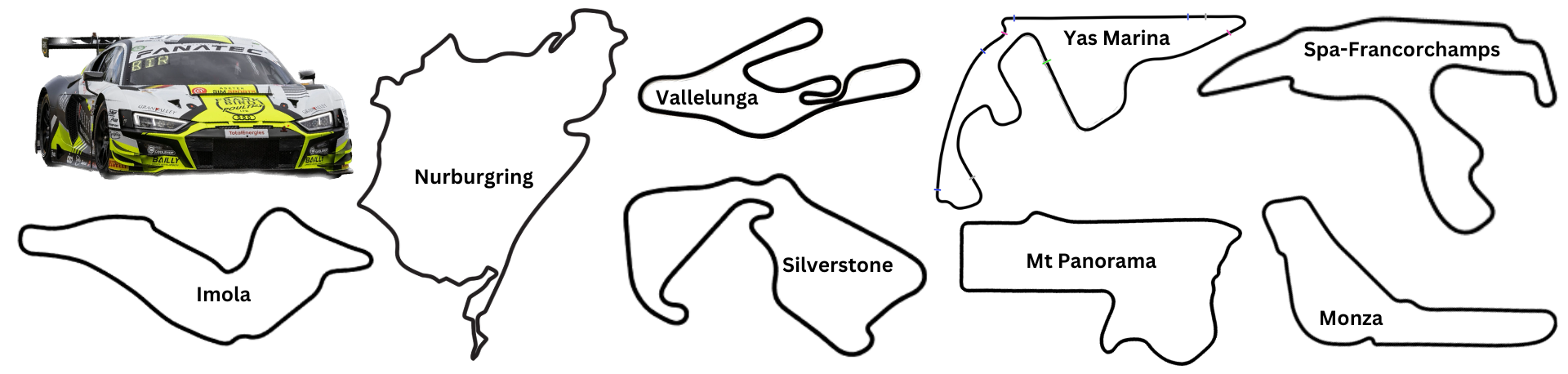}\\
        \end{tabular}
    }
    \caption{Vehicle and track layouts currently supported by AARK.}
    \label{fig:awesome_tracks}
\end{figure*}

    ACDG's initial release contains semantic mappings for eight circuits, shown in Figure \ref{fig:awesome_tracks}.
    These include six official AC maps: Imola, Circuit de Spa-Francorchamps, Autodromo Nazionale di Monza, Silverstone, Vallenlunga, Nordschleife, in addition to two community-made tracks: Mount Panorama and Yas Marina.
    Tracks were chosen based on their respective challenge to drivers and agents, alike, with varying levels of complexity.
    The complexity of a track ranges from Monza, which has many long straights and very few corners, to Nurburgring - one of the most challenging tracks in the world,  with 170 turns spaced over 25km.
    Mt Panorama and Spa feature large changes in elevation and blind corners. 
    Silverstone and Monza contain several corner elements that need to be connected to prioritise exit speed onto straights rather than speed through corners.
    Vallelunga focuses on extremely technical slow speed cornering with hairpin style loops.
    Yas Marina and Imola are contain a variety of circuit elements and are included to support parity with real-world autonomous racing leagues, such as A2RL~\cite{a2rl}.
    
    Each of these race tracks includes a semantic segmentation dataset and a pre-trained deep learning model for use in researchers' control solutions.
    If researchers wish to develop a dataset for a not yet supported circuit, a series of video tutorials demonstrating how to add semantic segmentation data generation support for new tracks are available here: \url{https://www.adelaideautonomous.racing/docs/acdg/getting-started/tutorials/}.

\begin{figure}
    \centering
    \includegraphics[width=0.46\textwidth]{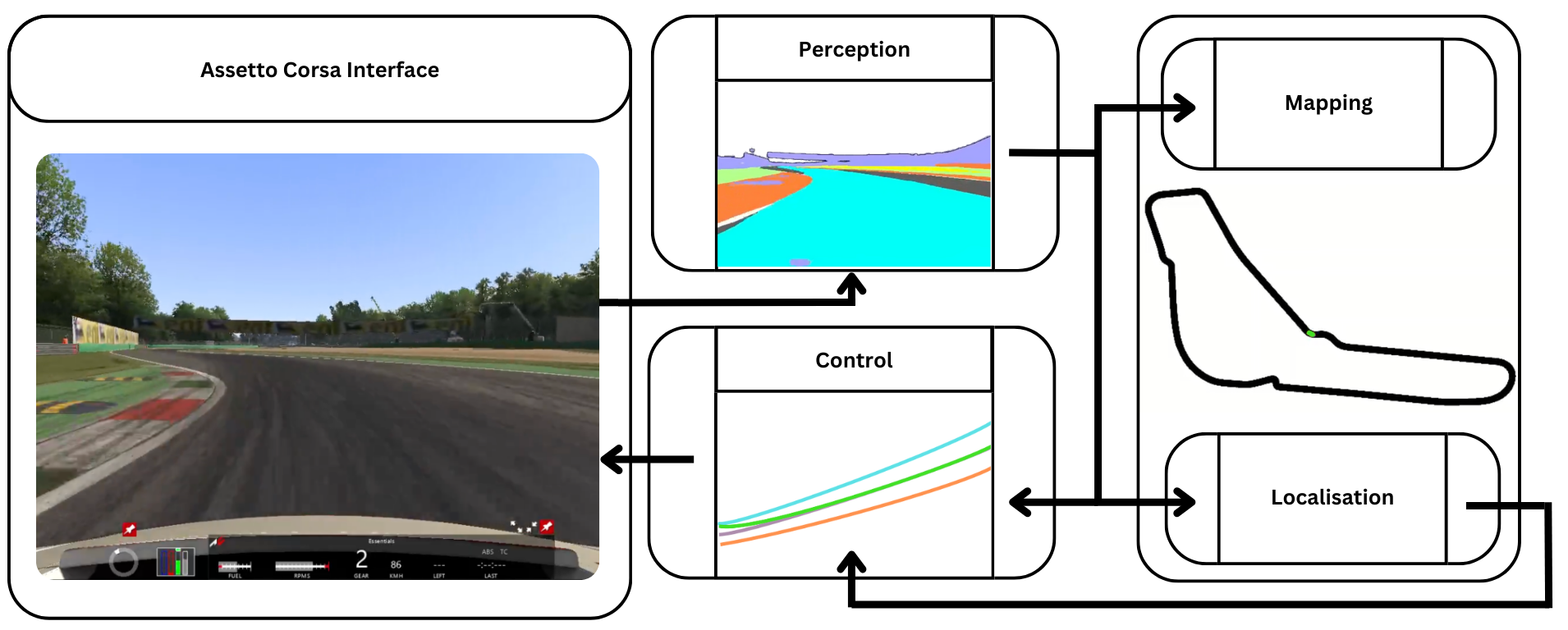}
    \caption{System diagram of ACMPC. Specifically how information from ACI is processed to produce vehicle control inputs.}
    \label{fig:acmpc}
\end{figure}

\subsection{DATASETS}
    To accelerate researcher's ability to participate in autonomous racing, a series datasets created using AARK are available here: \url{https://www.adelaideautonomous.racing/downloads/}.

\subsubsection{Behaviour}
    For each supported racetrack, associated recordings of human driving were gathered.
    Roughly eight laps of human control is available for use in both the ACI and MoTeC formats.
    MoTeC is compatible with i2, a free data analysis software used by professional motorsport teams to analyse driver and vehicle performance.
    By collecting telemetry data in this format, autonomously controlled vehicle behaviour can be compared against their human driver counter parts.
    This can provide insights into how autonomous control differs from race car drivers and vice versa.
    Recordings provided in the ACI format are compatible with behavioural cloning or supervised machine learning methods.
    Additionally, the ACI format can be used to generate a series of machine vision data for perception systems based on deep learning.

\subsubsection{Vision}
    The datasets used to train ACMPC's perception model are outlined in Table \ref{tab:segmentation-data}.
    These provide semantic, depth and normal map annotations for each of AARK's supported racetracks.
    Semantics consist of nine race track specific classes.
    Seven of these classes describe the type of surface: road, drivable, curb, carpet, grass, sand and water.
    Drivable is reserved to delineate the portion of road surface that is legal for vehicles to traverse while racing.
    A track limits label is included that allows models to perceive road markings which can indicate different features of importance around a racecourse.
    The remaining classes denote physical objects in the environment: people, vehicles, structures and vegetation.
    Structures encompass a super set of buildings, fences, safety barriers and signs.
    ACDG is capable of generating ground truth annotations for the following machine vision tasks: semantic segmentation, depth, normal maps, semantic LiDAR, and 3D object detection.
    Currently AARK provides pre-generated datasets for the first three of these tasks.

\section{FULL-STACK AUTONOMOUS CONTROL SOLUTION}

\subsection{ACMPC}
    Autonomous racing research carries with it a significant barrier to participation that can be daunting for researchers.
    On such barrier is the all-or-nothing nature of control solutions, requiring large amounts of upfront platform development before being able to experiment.
    To help overcome this, AARK includes a full-stack autonomous control solution.
    Commonly, an autonomous control stack consists of four key components: perception, localisation and mapping, planning, and control~\cite{VANBRUMMELEN2018384}.
    Although this pattern provides boundaries that separate out the concerns of each subsystem, it creates an all-or-nothing paradigm where all components must be present for an autonomous system to demonstrate competency.
    To overcome the upfront technical investment of designing, implementing and testing such a system, AARK provides the ACMPC package as a starting point for researchers.
    ACMPC is a full stack autonomous control solution capable of controlling vehicles around several circuits.
    This particular solution is an improved version of the controller that won the 2022 Learn 2 Race Challenge.
    A high level diagram of ACMPC is shown in Figure \ref{fig:acmpc}.
    ACMPC uses a semantic segmentation model to perceive the local track profile.
    These local precepts are used to map the circuit, localise the agent, and plan a trajectory using model predictive control (MPC).
    For a detailed explanation of the control solution, see our previous work: \textit{The Edge of Disaster: A Battle Between Autonomous Racing and Safety}~\cite{howe2022edge}.
    The specific implementation contained in ACMPC focuses on the independent and modular nature of each component in the autonomous control stack.
    This allows researchers to focus their efforts on a single piece of the system.
    The entire stack is written in Python and each component is contained in a separate process to allow performant parallel execution.
    A diagnostic dashboard which streams visualisations and status information while agents are driving is included to aid debugging and system analysis.
    A Twitch stream of ACMPC racing on a rotation of tracks can be found at \url{https://www.twitch.tv/vid3ojames}.
    This stream is a live real-time feed running on a desktop computer equipped with a Ryzen 5950X and RTX 4090.

\subsection{Evaluation}

\begin{table}
\centering
\caption{
Lap times achieved by ACMPC agents at different tracks in an Audi R8 LMS 2016.
Times are expressed as m:s.ms $\pm$ s.ms.
Results for multi-lap benchmarks show the average lap time plus minus one standard deviation.
}
\label{tab:acl2r-results}
\begin{tabular}{@{}lccc@{}}
\toprule
Track       &  \multicolumn{1}{c}{1-Lap} & \multicolumn{1}{c}{5-Lap} & \multicolumn{1}{c}{20-Lap} \\ \midrule
Monza  &  2:05.627  &  2:05.947 $\pm$ 0.345  &  2:06.229 $\pm$ 1.438 \\
Spa    &  3:19.021  & 3:19.333  $\pm$ 0.306 &  3:20.473 $\pm$ 1.182 \\
Silverstone & 2:28.568 & 2:28.728 $\pm$ 0.101 &  2:28.831 $\pm$ 0.167 \\
Vallelunga &  2:04.818 & 2:04.984 $\pm$ 0.125 & 2:05.271 $\pm$ 0.313 \\ \bottomrule
\end{tabular}
\end{table}

    Many existing benchmarks start vehicles from a standstill and focus on single lap pace~\cite{deepracer, learn_to_race}.
    Although this provides a common and easily controllable initialisation, it doesn't represent how drivers and vehicles are evaluated in reality.
    Vehicles with cold engines, tyres, and brakes are not in their optimal window of operation and are, therefore, unable to perform at their peak.
    Additionally, when vehicles start from a standstill, they do not receive the benefit of any speed they are carrying from a previous lap.
    In contrast, we choose to evaluate autonomous racing systems analogous to how humans are evaluated.
    Table \ref{tab:acl2r-results} shows the lap times achieved by ACMPC in three scenarios that examine different aspects of racecraft:
    
    \textbf{Single Lap Speed:} Agents are permitted any number of warm-up laps where they are able to bring the vehicle up to operational temperature and get a flying start.
    
    This enables improvement over agents that start from a standstill to those that use practice time to perform mapping, leverage the increased grip gained from managing their tyres, and carry speed from a previous lap into their hotlap.
    
    \textbf{Five Lap Consistency:} By extending the evaluation window from one to five consecutive laps, and averaging the achieved lap times, a better view of how well the agent can maintain race pace and its consistency as a driver can be discerned.
    
    \textbf{Twenty Lap Conservation:} Expanding the number of laps that are averaged, to a common length between pit-stops, further increases complexity.
    Over this time period, agents that learn to manage tyre temperatures, pressures, and degradation, while leveraging the decrease in platform mass due to fuel consumption, will outperform more naive solutions.

\begin{table}[]
\centering
\caption{
Lap times achieved by SAC agents at different tracks in an Audi R8 LMS 2016.
Times are expressed as m:s.ms $\pm$ s.ms.
Results for multi-lap benchmarks are not shown due to the SAC agents inability to achieve the required number of consecutive laps.
}
\label{tab:sac-results}
\begin{tabular}{@{}ccccccc@{}}
\toprule
       Monza & Spa      & Silverstone & Vallelunga & Bathurst & Imola  \\ \midrule
   1:58.072    & 2:31.457 & 2:23.979    &   1:38.765         & 2:12.865 &   1:53.304   \\ \bottomrule
\end{tabular}
\end{table}

\section{Reinforcement Learning}
\subsection{Soft Actor Critic}
        To demonstrate ACI's compatibility with reinforcement learning based control solutions, a gym interface and modified version of the Soft Actor Critic (SAC) agent implementation in ~\cite{remonda2024simulationbenchmarkautonomousracing} is provided as part of AARK.
        Although SAC may seem to achieve a better level of single lap control than ACMPC, there are significant differences in experimental conditions which prevent an apples to apples comparison of the two techniques.
        Most glaringly, the SAC agent, at both training and inference, receives a large amount of ground truth information from the simulation.
        This includes information such as the curvature of a reference racing line for the next 300 meters ahead of the vehicle's current position, the amount of slip each tyre is experiencing against the road surface, ego angular and axial velocities, and a collection of distance measurements from the vehicle's current position to the reference racing line and track limits.
        In contrast ACMPC executes vehicle control based on the current speed of the vehicle and a camera image.
        Experimentation on SAC agents under evaluation conditions were unable to produce the same level of consistency as ACMPC.
        SAC agents struggle to produce five consecutive laps without exceeding tracklimits or crashing, let alone the target duration of twenty.
        AARK's version of the SAC agent provided by~\cite{remonda2024simulationbenchmarkautonomousracing} makes significant optimisations to the procedures used for calculating state representations and training stability.
        The state calculating subroutines no longer requires a GPU to run in real time, significantly increasing the accessibility of the approach to those with less computational resourcing.
        Additionally, it was found that training stability was greatly improved by layer a curriculum based approach on top of SAC.
        The curriculum encourages agents to learn the racing line and complete laps before attempting to travel faster.
        This is done by limiting the reward received during training for going faster until an agent has completed a replay buffer's worth of successful episodes.
        From this point, reward for pace is increased each time a successful episode is completed.
        This process is repeated until reward derived from speed is effectively uncapped.
        Without this warm-up agent become stuck in a local reward maxima, where they fail to link a full lap together.
        Usually this manifests as a failure to learn to brake into tight corners, with agents greedily prioritising speed over positioning and longevity.
        A series of single lap baseline speeds for SAC agents trained on various circuits are reported in Table \ref{tab:sac-results}.

\begin{figure*}
    \centering
    \resizebox{\textwidth}{!}{
        \begin{tabular}{c}
        \includegraphics[width=0.3\textwidth]{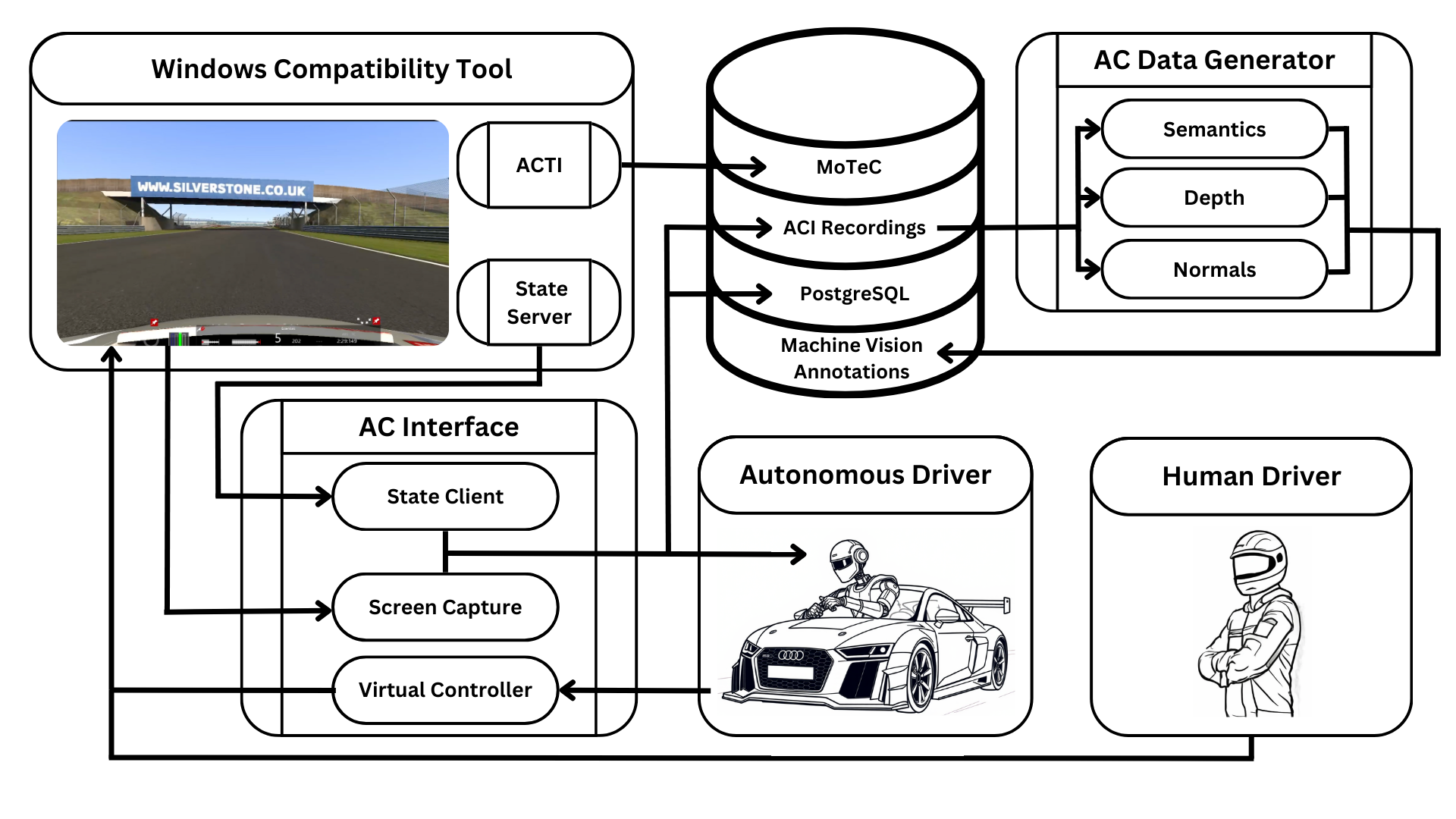}\\
        \end{tabular}
    }
    \caption{System diagram of ACI enabling interaction of autonomous agents with Asseetto Corsa.
    Recording of telemetry and machine learning data using ACI and ACDG is also shown for both autonomous and human drivers.}
    \label{fig:ac_interface}
\end{figure*}

\section{LIMITATIONS AND FUTURE WORK}

    In its current state, AARK supports one game, AC.
    However, AARK's implementation is as simulator agnostic as possible.
    Many popular racing games, such as Assetto Corsa Competizione and iRacing, expose real-time vehicle state information for use in motion platforms, steering feedback, ABS/TC vibrations, and custom dashboards.
    Although ACDG relies on AC specific features, an agent trained on ACDG generated data can be tested on other simulators by adapting ACI such that frames are captured from a different application window and state information is collected accordingly.
    Therefore, as new racing simulators are released, autonomous control solutions developed using AARK can continue to be evaluated and compared on new simulators into the future.
    
    Support for multiple autonomous agents competing against one another in online lobbies, ROS integration, longer format races with pit stops, more circuits, and different motorsport formats such as autocross and rally are all on AARK's development road map.
    Enabling real-time emulation of LiDAR and production of ground truth machine vision annotations, by moving ray casting operations to a co-processor, is another key milestone.
    These additional features will further expand the types of research that can be done using AARK.

    Finally, the simulation to real gap that exists between any digital twin and reality has not been fully evaluated in the case of Assetto Corsa and reality.
    In the future it would be possible to explore this explicitly if provided with access to an autonomous racing platform and circuit.

\section{CONCLUSION}
    AARK lays a foundation for autonomous racing research that presents many exciting and diverse opportunities.
    In our research, AARK provides us the latitude to explore advanced mapping, localisation, and control techniques that require increased simulator fidelity.
    For the wider community, AARK provides an accessible, realistic, and customisable autonomous racing agent development kit bridging the gap between reinforcement learning and classical system design philosophies.
    By providing an open-source, modular, full stack autonomous system in the form of ACMPC, researchers can get started immediately by branching off from a working system.
    AC's realism and extensive modification community make it the perfect starting simulator, ensuring adaptability to accommodate many different types of approaches to autonomous racing systems.
    Finally, AARK is designed as a generic simulation interface that makes swapping out components, such as AC for another simulator, possible.
    Ultimately, AARK allows researchers to get closer, sooner, to demonstrating capable autonomous system to explore algorithmic approaches and justify investment into their autonomous programs prior to the acquisition of physical platforms or proprietary commercial simulators.

\section*{APPENDIX}


\subsection{Implementation details of ACI}\label{app:aci_implementation}
    To allow an autonomous agent to interact with AC, ACI provides bundled captured frames with state information from the simulator to autonomous systems and allows those systems to input vehicle controls in response.
    As AARK has a strong emphasis on accessibility it was important to ensure execution of simulation and autonomous systems are as flexible as possible.
    For example, if a researcher prefers Linux operating systems ACI enables the programmatic execution and configuration of AC on Linux machines via the use of compatibility tools.
    So far, support for using Cross Over and Steam Proton are included in ACI.
    These compatibility tools create a separate environment for execution of the Windows native applications and therefore cannot be directly accessed by the Linux host.
    This air gap between the host and the simulator is bridged by ACI's state server, which packages raw state information and forwards it to a client on the host via a socket.
    In this configuration, where both simulator and autonomous system are co-located on a host machine, an extremely low latency between observations, processing and control is achieved.
    Equally, this approach to implementation allows for a network connection between the simulator and autonomous system; enabling exploration of hardware in the loop or distributed system based approaches.
    Furthermore, ACI provides bindings which allow an autonomous system to be executed inside a docker container, vastly improving the development experience and ease of reproducibility.
    Figure~\ref{fig:ac_interface}, shows a detailed overview of the main components of ACI and ACDG indicating their relationship to one another.
    For specific details and usage visit AARK's documentation website here: \url{https://www.adelaideautonomous.racing/docs-aci/}

\subsection{Included for convenience}
    Included with AARK are several quality of life enhancements that have made development and tuning of our autonomous racing agent more intuitive and convenient. 
    These internal tools are provided to the community to help researchers get started in such a complex field.
    ACMPC includes a standalone PyQt diagnostics dashboard/system monitor, a non-interactive version of which can be seen on the left of Figure \ref{fig:aci}.
    This dashboard is capable of visualising system behaviours such as localisation, segmentation, and planning.
    Visualisations are computed in a separate processes from the autonomous system itself, allowing for non-blocking execution of critical control processes; minimising visualisations impact on system performance.
    In a separate repository, code for training and evaluation semantic segmentation models on datasets produced by ACDG is provided.
    This allows researchers to reproduce reported results from scratch, improve on or experiment with approaches to modeling and train on new data generated for novel circuits.
    An additional telemetry format is supported by ACI to allow convenient storage and access by data scientists.
    ACI does this by providing a telemetry writer that stores simulation state information in a PostgreSQL database.
    This allows interaction and manipulation of telemetry data via SQL which has been useful for creating custom analysis metrics or windows for those less familiar with MoTeC.
    Lastly, AC-Extras (ACE) is able to read and assemble extracted simulation vehicle data into Python objects is included.
    This provides programmatic access to vehicle information such as wheelbase and suspension geometry for all simulated vehicles in AC.
    ACE is also packaged with a series of mathematical convenience functions which use parsed vehicle geometry information to calculate transformations between steering inputs and tyre angles.

\section*{ACKNOWLEDGMENTS}
This research was supported by PhD Studentship via Australian Government Research Training Program (RTP) Scholarships awarded to JPB and MRH.
Lockheed Martin Australia supported this research via research scholarships awarded to JPB and MRH.

\addtolength{\textheight}{-12cm}   

\bibliographystyle{IEEEtran}
\bibliography{references}

\end{document}